\documentclass[10pt]{article}

\usepackage[margin=1in]{geometry}
\usepackage[T1]{fontenc}
\usepackage[utf8]{inputenc}
\usepackage{amsmath}
\usepackage{amssymb}
\usepackage{booktabs}
\usepackage{graphicx}
\usepackage{microtype}
\usepackage[round]{natbib}
\usepackage{xcolor}
\usepackage{hyperref}

\hypersetup{
  colorlinks=true,
  linkcolor=blue!55!black,
  citecolor=blue!55!black,
  urlcolor=blue!55!black,
  pdftitle={Correct Is Not Governed: Provenance Integrity in Agentic Workflows},
  pdfauthor={Jesus Salas (jesus.salas@gmail.com), Independent Researcher},
  pdfsubject={Provenance integrity and governed execution in agentic workflows}
}

\newcommand{\matrixsys}{\textsc{Matrix}}

\title{Correct Is Not Governed:\\
Provenance Integrity in Agentic Workflows}
\author{Jesus Salas\thanks{The author is employed by Microsoft. This work was
performed independently, outside the scope of that employment, on personal time
and equipment. It does not represent the views or positions of Microsoft
Corporation.}\\
Independent Researcher\\
\texttt{jesus.salas@gmail.com}}
\date{August 2026}

\begin{document}

\maketitle

\begin{abstract}
Agentic workflows are commonly evaluated by whether they reach the correct
outcome. That is insufficient in institutional settings, where a correct action
may rely on the wrong authority, an unsupported completion claim, or work made
stale by a later change. We define \emph{governed execution} as work whose
decisions, completion, and response to change are supported by inspectable
provenance. We present \matrixsys, a deterministic causal-state layer that
records authority and fact dependencies, verifies completion evidence, and
selectively invalidates affected work. Across controlled comparisons, governed
and direct workflows often reached the same outcomes, but only the governed
path consistently preserved governing evidence, refused unsupported closure,
and limited recovery to dependent tasks. A role-separated transfer challenge
then failed: a deterministically enforced completeness contract severely
over-blocked synthetic packets produced outside its authoring context. These
results do not establish \matrixsys{} as a general accuracy enhancer; they
support its primary role as an institutional integrity layer for making agentic
work auditable and independently verifiable.
\end{abstract}

\section{Introduction}

Correct execution is not necessarily governed execution. An agent can choose
the expected action while relying on a draft, expired waiver, or adjacent-domain
exception. It can announce that a task is complete without producing the
evidence required by the task's definition of done. It can also reach a current
business outcome after a policy change while leaving no account of which prior
work became stale, which evidence remained valid, or why only some tasks were
repeated. Ordinary success metrics can score each workflow as correct.

These failures concern the institutional account of the work rather than the
surface answer alone. An enterprise reviewer may need to establish what
authority and facts produced a decision, whether execution actually satisfied
its declared obligations, and how later changes propagated through prior work.
A transcript, retrieved document set, or agent assertion can provide evidence
for that account, but none is automatically the governing account itself.
This concern is also reflected in institutional guidance. Article 12 of the
European Union AI Act requires logging capabilities for high-risk AI systems to
support traceability, while the NIST AI Risk Management Framework emphasizes
documentation across governance and risk-management activities
\citep{eu2024aiact,nist2023airmf}. These sources do not prescribe the
architecture studied here, and \matrixsys{} is not claimed to establish
regulatory compliance. They motivate treating reconstructable evidence as a
system requirement rather than a presentation feature.

We organize this problem around three forms of provenance integrity:

\begin{description}
  \item[Decision provenance] identifies the current authority and accepted
  facts that materially governed an action, while distinguishing them from
  merely retrieved or semantically related evidence.
  \item[Execution provenance] connects each obligation to independently
  checkable completion evidence and prevents an agent's own completion claim
  from serving as sufficient proof.
  \item[Change provenance] records which decisions, tasks, receipts, and prior
  verdicts depend on a superseded authority or fact so that recovery can
  invalidate stale work without discarding unaffected proof.
\end{description}

We investigate whether these properties can be externalized into a
deterministic layer around probabilistic agents. \matrixsys{} records typed
operations over versioned authority, facts, tasks, completion receipts,
dependencies, invalidations, replacements, and supervisory verdicts. Models
continue to interpret documents, propose plans, and perform open-ended work.
The framework owns identity, transition validity, mechanically declared
preconditions, receipt matching, bounded rework, and replay. This separation is
intended to make the lifecycle inspectable without pretending that governance
makes semantic judgment infallible.

The evaluation deliberately includes cases where business outcomes tie. In an
authored retrieval corpus, direct RAG and governed paths selected every expected
action, but differed in whether non-governing records were cited. In matched
completion fixtures, the same Phi-4 plan and a generic self-review preceded
both conditions; the direct control nevertheless self-closed on unsupported
execution claims, while \matrixsys{} preserved residual work and required
valid receipts. In a versioned-change fixture, both conditions stayed current,
but dependency lineage reduced task reexecution from 18 to 3 across three
seeds. A separate metadata-decidable invariant forced escalation when a required
authority class was absent, even though Phi never proposed escalation itself.

These results motivate the paper's central claim:

\begin{center}
\fcolorbox{black!55}{black!2}{%
\parbox{0.9\linewidth}{%
\centering
\textbf{Task correctness and governed execution are different properties.}\par
\smallskip
A workflow is governed only when the authority behind its decisions, the
evidence behind its completion, and the effects of later change are explicit
and independently checkable.}}
\end{center}

The distinction changes what counts as a useful result. A tied business outcome
can still reveal a provenance or assurance difference. Conversely, a
deterministic invariant can perfectly implement an authored contract while the
contract itself fails to transfer. We therefore report a failed synthetic
contract-transfer test alongside the positive mechanism results.

\paragraph{Contributions.}
This paper makes three contributions:

\begin{enumerate}
  \item It defines governed execution as a conjunction of decision, execution,
  and change provenance, and makes each property measurable separately from
  task correctness.
  \item It reports a controlling negative result: a deterministically enforced
  completeness contract achieved perfect sensitivity but zero specificity on
  the initial role-separated transfer set, showing that mechanism fidelity does
  not establish contract validity.
  \item It presents an implemented lifecycle joining versioned authority and
  facts to decision admission, task issuance, receipt-verified closure,
  supervisory verdicts, and dependency-scoped recovery, with controlled
  same-outcome demonstrations across those surfaces.
\end{enumerate}

The contribution is this lifecycle-wide governed-work formulation and its
matched evaluation, not the novelty of append-only logs, runtime contracts,
citation checking, or dependency replay in isolation. Section~8 compares those
established and concurrent mechanisms directly.

\paragraph{Scope.}
The principal studies use small, self-authored synthetic fixtures and repeated
model seeds. They establish implemented mechanisms, not enterprise prevalence,
human-review cost, semantic completeness, or general accuracy gains. The paper
does not claim superiority over citation-verifying RAG, objective truth from
ledger acceptance, natural detection of every relevant change, or validated
process forecasting. The next claim-bearing gate is an independently authored,
human-reviewed governed-work task with operational-overhead and false-block
measurement; additional self-authored rungs cannot establish that validity.

\section{Operational Model of Governed Execution}

\subsection{Lifecycle state}

At time $t$, the workflow state contains six first-class sets: versioned
authority $A_t$; accepted facts $F_t$; admitted decisions $Q_t$; issued tasks
and their definitions of done $T_t$; completion receipts $R_t$; and supervisory
verdicts $V_t$. Typed dependency edges record which versions were read or
produced by each later object.

This state is an institutional claim, not an objective-truth oracle. Authority
can be incomplete, facts can be wrong, and models can misinterpret evidence.
The purpose of the state is to make those accepted claims explicit, versioned,
and revisable.

\subsection{Decision provenance}

Let $\mathcal{K}$ be the available institutional corpus, $D_K \subseteq
\mathcal{K}$ the top-$K$ records retrieved for a decision, and $G \subseteq
\mathcal{K}$ the records that are effective, authoritative, and applicable.
Retrieval recall asks whether every required member of $G$ appears in $D_K$.
It does not ask whether the system avoids promoting records in $D_K \setminus
G$ into governing state.

If $C$ is the set cited as materially governing the action, citation soundness
requires $C \subseteq G$. When every member of $G$ is required to justify the
decision, completeness also requires $G \subseteq C$. A decision may therefore
select the expected action while remaining provenance-incorrect.

\subsection{Execution provenance}

For a task $\tau \in T_t$, let $E_\tau$ be the current evidence attached to its
completion receipt and $\operatorname{DoD}_\tau$ its declared definition of
done. The task is complete only when an independent verifier accepts
$E_\tau$ against $\operatorname{DoD}_\tau$ under the current authority and fact
state. The executing agent's statement that it completed $\tau$ is an input
claim, not sufficient evidence. A failed receipt remains in history; a
successful retry creates a later receipt rather than overwriting the failed
attempt.

\subsection{Change provenance}

When authority or fact version $x^v$ is superseded, the affected set contains
the current decisions, tasks, receipts, and verdicts whose dependency paths
reach that version. A governed recovery invalidates those dependent objects,
preserves current objects outside the affected set, issues replacement
obligations, and records how new objects supersede stale ones. Broad reruns can
also restore currentness, but they do not demonstrate dependency-scoped
recovery.

\subsection{Correctness and governance are measured separately}

We use \emph{business correctness} for whether a workflow reaches its expected
business outcome. We use \emph{governed execution} for the conjunction of sound
decision provenance, verified execution provenance, and current change
provenance. This is an operational definition, not a theorem.

The empirical separation criterion is a matched pair in which both workflows
reach the expected business outcome but differ on at least one provenance
property. The studies test whether such pairs occur under controlled
conditions. Deployed work requires both correctness and governance; the claim
is that one cannot be used as a measurement proxy for the other.

\subsection{Research questions}

\begin{description}
  \item[RQ1:] When direct RAG and governed retrieval select the same expected
  actions, do they differ in which evidence is cited or promoted as governing?
  \item[RQ2:] Can deterministic receipt verification distinguish model-claimed
  completion from evidence-backed completion after a generic self-review?
  \item[RQ3:] When both conditions observe an authority change and remain
  current, can declared dependency lineage reduce recovery churn without stale
  residue?
  \item[RQ4:] Can a metadata-decidable completeness invariant force appropriate
  refusal without degenerating into refusal on every complete packet?
  \item[RQ5:] Can the selected decision, execution, change, and refusal controls
  compose in one bounded workflow while preserving their component checks?
\end{description}

\section{The \matrixsys{} Substrate}

\subsection{Design boundary}

\matrixsys{} separates deterministic institutional mechanics from
model-dependent cognition. Models interpret documents, judge applicability,
propose plans, perform work, and suggest repairs. The framework owns stable
identity, direct references, transition validity, materialization, declared
preconditions, receipt matching, dependency traversal, bounded retry counters,
and replay.

This boundary does not make semantic judgment deterministic. If a model admits
the wrong policy as applicable, the ledger can preserve that wrong judgment
faithfully. The deterministic claim is narrower: once authority, facts,
obligations, and evidence contracts are declared, the runtime can enforce the
associated state-transition rules independently of a model's willingness to
self-enforce them.

\subsection{State model}

The implemented prototype exposes the six logical sets from Section~2 through
an append-only event log and deterministic materialization
(Table~\ref{tab:system-components}).

\begin{table}[t]
\centering
\small
\begin{tabular}{p{0.24\linewidth} p{0.67\linewidth}}
\toprule
Surface & Recorded responsibility \\
\midrule
AuthoritySet & Policies, delegations, status, applicability metadata,
versions, and supersession. \\
FactSet & Accepted case facts with source references, versions, explicit
unknowns, and conflicts. \\
DecisionSet & Admitted rulings, selected authority and facts, dependencies,
revisions, and status. \\
TaskSet & Obligations with owner, dependencies, preconditions, permitted
actions, required outputs, and definition of done. \\
CompletionReceiptSet & Immutable attempts, tool or external evidence,
timestamps, task references, and verification status. \\
SupervisorVerdict & Completed, rework-required, blocked, rejected, or
escalated status with residual obligations. \\
Event and dependency state & Ordered operations, read/produce edges,
invalidations, replacements, and replay. \\
\bottomrule
\end{tabular}
\caption{Logical governed-work surfaces in the current prototype.}
\label{tab:system-components}
\end{table}

\begin{figure}[t]
\centering
\includegraphics[width=\linewidth]{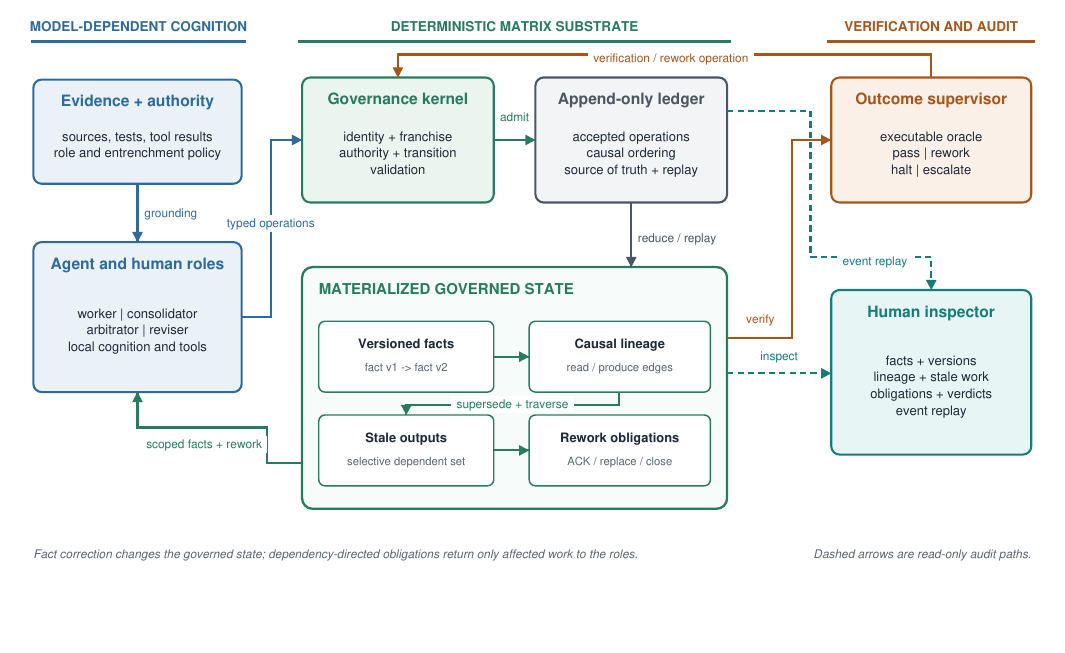}
\caption{Implemented causal-state architecture. Model and human roles submit
typed proposals; the governance kernel validates state transitions and records
them in an append-only ledger. Materialized facts, lineage, stale work, and
rework obligations feed an outcome supervisor, while the inspector exposes
both current state and replayable history. Tasks, receipts, and verdicts use
the same event and dependency path.}
\label{fig:architecture}
\end{figure}

\subsection{Governed-work lifecycle}

A complete run follows the same ordered lifecycle across the product fixtures:

\begin{enumerate}
  \item record applicable authority and case facts, including explicit
  unknowns;
  \item accept or reject a model-proposed decision and obligation plan;
  \item issue each task only after its declared preconditions are satisfied;
  \item preserve every execution attempt and attach evidence to a receipt;
  \item verify each receipt against the task's definition of done;
  \item emit a supervisor verdict with any residual obligations;
  \item request bounded, targeted rework or escalate when the retry budget is
  exhausted; and
  \item on authority or fact change, invalidate dependent state and repeat only
  the required portion of the lifecycle.
\end{enumerate}

The event log is append-only. A retry does not erase a failed receipt, and a
replacement task does not erase the authority version under which its
predecessor ran. Current materialized state and historical reconstruction are
therefore separate views over the same operations.

\subsection{Decision admission}

For corpus-backed decisions, a model adjudicator receives the task and the
retrieved records with their public identity, authority, status, and content.
It returns a resolution and the record ids it judges applicable. The kernel
checks that selected ids came from the recorded search and that later fact
acceptance follows valid transitions. Accepted records become explicit facts
and are bound to downstream work.

The kernel does not compare the model's selection with private applicability
labels. Any retrieved record selected by the adjudicator can be promoted, and
an incorrect promotion is scored only after execution. The mandatory-governance
errors in Experiment 5 confirm that this semantic step remains fallible.

For plan admission, a contract can require specific authority or fact
references without mutating model output. A failed proposal receives an exact
residual obligation and a bounded opportunity to return a replacement. Only a
replacement that independently passes the same contract is admitted.

\subsection{Receipt verification and bounded rework}

An execution adapter emits a claim and evidence. The supervisor evaluates that
evidence against the task-specific definition of done. Invalid evidence creates
a \texttt{rework\_required} verdict and preserves the task as residual. A later
attempt creates a new receipt and is independently checked. On every initial or
revised receipt set, the current kernel recomputes the effective verdict from
the contract checks and rejects a proposed verdict that contradicts those
checks or their residual obligations. Model completion language cannot authorize
closure. Exhausting the declared retry allowance produces
\texttt{escalated}; it does not silently convert the last model claim into
completion.

Definitions of done are domain contracts supplied to the framework. Matrix can
deterministically implement a contract, but it cannot establish that the
contract captures every semantic requirement of the real process.

\subsection{Version change and selective recovery}

When a new authority or fact version is admitted, the runtime records
supersession, traverses declared read dependencies, marks affected tasks,
receipts, and prior verdicts stale, and creates replacement obligations. State
outside the transitive dependent set remains current. A conservative full
rerun is a safe alternative when dependency lineage is unavailable; the C6
comparison measures the work avoided by the declared graph, not a correctness
advantage over that control.

\subsection{Structurally decidable refusal}

A public completeness contract may declare mechanically required authority
classes, domains, multiplicity, or applicability tags. If the visible candidate
set lacks a required class, the kernel records the model proposal but blocks a
confident ruling and makes escalation effective. Private answer labels are not
consulted.

This guarantee is intentionally bounded. It can decide that a declared source
class is absent; it cannot decide whether an available source semantically
covers an edge case, whether the contract omitted a requirement, or whether an
unknown open-world policy exists.

\section{Experimental Design}

\subsection{Evidence selection and reporting}

The experiments preceded the present three-provenance framing. We therefore
freeze a claim-evidence map before this rewrite and apply three reporting rules.
First, studies are selected because they isolate decision, execution, or change
provenance, not because \matrixsys{} wins on task accuracy. Second, original
scenarios, seeds, controls, repairs, and claim boundaries remain unchanged.
Third, counts from different studies are not pooled: scenarios, repeated seeds,
and deterministic action assertions represent different units.

Table~\ref{tab:study-map} maps every evidence source reported in the paper.
Studies 1--5 address the five research questions. Study 6 is a controlling
transfer failure rather than favorable downstream evidence. L4 is retained as
a matched decision-admission walkthrough and Experiment 3 as a supporting
state-surface stress test. The table maps paper labels to internal artifact
labels so the claims remain traceable to frozen reports.

\begin{table}[t]
\centering
\small
\begin{tabular}{p{0.12\linewidth} p{0.22\linewidth} p{0.27\linewidth} p{0.28\linewidth}}
\toprule
Paper label & Property & Comparator & Primary unit \\
\midrule
Study 1 (Exp. 5) & Decision provenance & Competent direct RAG & Five scenarios $\times$ three seeds \\
L4 demo & Decision admission & Shared initial Phi proposal; direct control & Three matched seeds \\
Study 2 (C4--C5) & Execution provenance & Same Phi plan plus generic self-review & Three matched seeds per rung \\
Study 3 (C6) & Change provenance & Conservative full rerun & Three matched seeds \\
Study 4 (Exp. 6) & Structural refusal & Raw model proposal and complete contracts & 12 incomplete and 120 complete executions \\
Study 5 (C7) & Composition & Same self-review control plus Phi compaction & Three matched seeds \\
Support (Exp. 3) & State-surface stress & Phi-compacted summary & Three cells $\times$ 40 paired seeds \\
Study 6 (Exp. 7) & Contract transfer & Reviewed packet completeness & Thirty synthetic packets \\
\bottomrule
\end{tabular}
\caption{Evaluation map. Repeated seeds are stability checks, not independent
task samples; action assertions are implementation checks.}
\label{tab:study-map}
\end{table}

\paragraph{Artifact scope.}
The accompanying reproducibility artifact contains the prototype source and
tests, scenario and contract inputs, frozen machine-readable run reports,
analysis scripts, the claim-evidence map used for this rewrite, and build
instructions. Repeated and repaired diagnostics remain labeled separately from
claim-bearing artifacts. A stable archive identifier is a release requirement,
not evidence supplied by the current manuscript.

\subsection{Study 1: decision provenance after retrieval}

Experiment 5 contains five public scenarios and 13 authored institutional
records. Three scenarios have one effective governing policy. The remaining
two have no governing record: a harmless internal theme change and an
unresolved high-impact biometric proposal. Distractors include drafts, expired
waivers, research notes, adjacent-domain policies, and narrow exceptions.

Expected actions, resolutions, and governing ids reside in a private control
file used only for scoring and scripted qualification. Automated payload scans
reject private control keys. Four conditions are compared:

\begin{enumerate}
  \item fixed scope, which retains the initial proposal without retrieval;
  \item direct RAG, which uses the shared query and top-$K$ records to emit an
  action and governing citations;
  \item model-signaled governance, which retrieves and adjudicates only after a
  typed scope interrupt and then binds accepted facts before revision; and
  \item mandatory governance, which runs the same adjudication and binding path
  for every scenario.
\end{enumerate}

All retrieval conditions use local Phi-4 Q4, three seeds, temperature 0.2, a
4,096-token role budget, BM25-style retrieval, and $K=4$. The query planner is
physically shared. Direct RAG and the governed reviser receive the same action
contract and authority/status metadata. Both are explicitly told to reject
drafts, expired waivers, research notes, unrelated jurisdictions, and narrow
exceptions. Governed paths add an applicability call, so inference count is not
matched. The condition-call totals exclude the shared 15-call query planner:
direct RAG uses 15 action calls, model-signaled governance uses 12 applicability
and nine action-revision calls, and mandatory governance uses 15 applicability
and 11 action-revision calls.

Primary clean success requires the expected action with no unsafe completion,
unsupported citation or promotion, false halt, unnecessary rework, or missing
required grounding. Action correctness and provenance integrity are also
reported separately.

\subsection{Study 2: claimed versus verified completion}

C4 and C5 are cumulative vendor-onboarding fixtures. Both conditions use the
same local Phi-4 Q4 checkpoint, temperature zero, and seeds 1--3. Both
receive the same initial plan and one active, oracle-free generic self-review.
The checklist asks the model to represent each applicable obligation exactly
once, remove inapplicable work, cite governing authority and applicability
facts, name independently checkable evidence, and mark the plan ready only when
all work is represented and verifiable. It contains no case-specific answer.

After self-review, the direct control executes its revised plan and accepts its
own completion claims. \matrixsys{} independently compiles obligations, checks
preconditions, issues tasks, validates receipts against obligation-specific
definitions of done, and closes only from a supervisor verdict.

C4 gives both arms the same exact six-obligation plan and six execution claims.
One DPA claim lacks its contract-required evidence. Correct governed behavior
is \texttt{rework\_required} with the DPA obligation preserved as residual; C4
does not include a successful recovery.

C5 adds one receipt-targeted retry. The first DPA attempt must remain in history,
the correction must name the missing evidence, the retry must be independently
validated, and closure may follow only from a complete current receipt set. A
deterministic regression repeats invalid evidence through the bounded retry
allowance and expects escalation. That branch tests implementation behavior,
not model failure frequency.

Primary measures are premature closure, evidence-backed completion, residual
obligations, bounded rework, supervisor verdict, and exact lifecycle-action
assertions.

\subsection{Study 3: change provenance and recovery scope}

C6 starts from an exact six-obligation plan and completed initial execution.
Before closure, privacy authority v2 supersedes v1 and changes only the DPA
evidence contract. Both conditions receive the amendment. \matrixsys{} follows
declared dependencies and reissues only DPA. The strong direct control lacks
dependency lineage and conservatively reruns all six tasks, ensuring that it is
current rather than stale.

Primary measures are current lifecycle exactness, stale residue, affected task
reissues, preserved unaffected receipts, and action assertions. Because both
conditions are designed to be safe, task accuracy is expected to tie; the
measured difference is recovery scope.

\subsection{Study 4: metadata-decidable refusal}

Experiment 6 contains 30 public policy sources, 12 standard scenarios across
six conflict families, and two missing-authority controls. Each control removes
a source class required by a public completeness contract. The contract is
visible to the framework but never enters the model prompt; private expected
actions and source labels are not consulted by enforcement.

On the incomplete controls, the framework records the raw adjudicator proposal,
checks whether the required authority class exists, blocks a confident ruling
when it does not, and makes escalation effective. The three-seed qualification
contains 12 such executions across two controls and two primary packet surfaces.

After observing sensitivity, complete contracts were added for all 12 standard
scenarios to test the degenerate strategy of escalating everything. The frozen
five-seed primary pilot supplies 120 complete executions across two packet
surfaces. This specificity amendment is post hoc and uses the same authored
fixture. We report both repeated executions and unique contract counts.

Ordinary policy-conflict reasoning remains model-dependent and is not a primary
result here. The five-seed graph condition achieved 27/60 exact adjudications
versus 21/60 for a matched flat packet, with descriptive paired McNemar
$p=0.238$. We treat that difference as directional and unpowered.

\subsection{Study 5: composition under bounded history}

C7 composes conditional selection, an unknown-gated dependency, bounded receipt
rework, selective authority recovery, and supervisor closure. The estimated raw
history is 6,160 tokens against a frozen 4,096-token treatment budget. Phi
performs both the fair control compaction and all closure reviews. Three probes
receive fresh truth, mechanically scoped Matrix state, or the Phi-compacted
control. The composition gate requires the governed lifecycle, every component
action, and all budget guards to pass; it does not require the compactor to fail.

\subsection{Supporting lifecycle and state-surface tests}

L4 is the fourth rung of a synthetic vendor-onboarding lifecycle ladder and
supplies an interpretable decision-admission walkthrough. It uses a matched
governed-versus-direct design across three Phi seeds. Both arms share the
initial Phi proposal and diverge only afterward: the direct arm executes without
governed admission, while \matrixsys{} may issue bounded decision rework before
execution. The initial proposal omitted a required preferred-vendor fact
reference in all three seeds. \matrixsys{} rejected the proposal, issued the
exact residual, allowed one model revision, and revalidated the replacement
before any task execution. The direct path retained the omission, while both
paths ultimately reached the expected business state. We use L4 to illustrate
the lifecycle rather than as a separate population estimate.

Experiment 3 remains a supporting test of whether provenance state can become
outcome-relevant under forced context loss. A 25,254-token two-correction history
exceeds a 6,144-token budget. The same frontier reviser receives either a
model-generated transcript summary or an approximately 5,579-token mechanical
scope query containing answer and distractor lineage. Three preselected
compaction cells use 40 paired seeds. A faithful prose rendering of the same
scope query tests whether record syntax is the active ingredient. This study is
reported as a controlled state-surface stress test, not a general product
accuracy claim.

\subsection{Study 6: transfer challenge}

Experiment 7 asks whether an authored structural completeness contract transfers
to independently contextualized synthetic packets. Role-separated agent
contexts curated 30 packets; two fresh domain-reviewer contexts and a later
adjudicator assessed completeness without access to the machine contract. This
is procedural information separation on a shared filesystem, not independent
human expertise or a security boundary.

The controlling gate requires both incomplete sensitivity and complete
specificity. If the gate fails, downstream workflow comparisons remain
diagnostic. A later semantic-normalization pass uses a separately expanded,
balanced 40-packet diagnostic set (20 reviewer-complete and 20
reviewer-incomplete) assembled after the original 30-packet corpus was exposed.
It is therefore labeled a burned-corpus diagnostic rather than a new transfer
test.

\section{Results}

The studies answer different questions and use different units, so we do not
report a pooled Matrix success rate. The recurring pattern is instead whether
task outcome and provenance integrity agree or diverge.

\subsection{Study 1: equal actions, different governing evidence}

Table~\ref{tab:decision-provenance} reports Experiment 5. Every retrieval
condition selected the expected action in all 15 repeated runs. Direct RAG was
therefore perfect on action accuracy, yet it cited non-governing records in six
runs. The model-signaled governed path cited or promoted none. Mandatory
governance promoted and cited an irrelevant exception in two runs.

\begin{table}[t]
\centering
\small
\begin{tabular}{lcccc}
\toprule
Metric & Fixed & Direct RAG & Signaled gov. & Mandatory gov. \\
\midrule
Expected action & 3/15 & 15/15 & 15/15 & 15/15 \\
Primary clean success & 3/15 & 9/15 & 15/15 & 13/15 \\
Unsafe completion & 12/15 & 0/15 & 0/15 & 0/15 \\
Unsupported citation & 0/15 & 6/15 & 0/15 & 2/15 \\
Unsupported promotion & 0/15 & 0/15 & 0/15 & 2/15 \\
Applicability + action calls & $0+0$ & $0+15$ & $12+9$ & $15+11$ \\
\bottomrule
\end{tabular}
\caption{Study 1 (Experiment 5) over five authored scenarios and three seeds.
Primary clean success combines action, safety, and provenance requirements.
Call counts exclude the shared 15-call query planner.}
\label{tab:decision-provenance}
\end{table}

Relative to direct RAG's 15 condition-specific calls, model-signaled governance
used 21 calls ($1.40\times$) and mandatory governance used 26
($1.73\times$). Including the shared 15-call query planner, the corresponding
end-to-end totals were 30, 36 ($1.20\times$), and 41 ($1.37\times$). These are
model-call counts, not measurements of tokens, latency, or monetary cost.

The six direct-RAG errors were stable by scenario: in every seed it cited an
internal-tool exception and public-brand standard for the harmless theme task,
and a non-authoritative research note for the unresolved biometric task. The
shared planner interrupted all 12 consequential cases and none of three
harmless controls; retrieval returned every required governing record in the
nine applicable-policy runs. The observed errors were therefore not failures to
make the relevant policy available.

Clean governed grounding was not answer-key-constrained. The adjudicator could
select any retrieved id, and the framework would bind that choice. The two
mandatory-condition promotions show that semantic adjudication remained
fallible. The result supports staged adjudication plus governed binding versus
the measured direct-RAG pipeline, not governance by construction.

\begin{figure}[t]
\centering
\includegraphics[width=\linewidth]{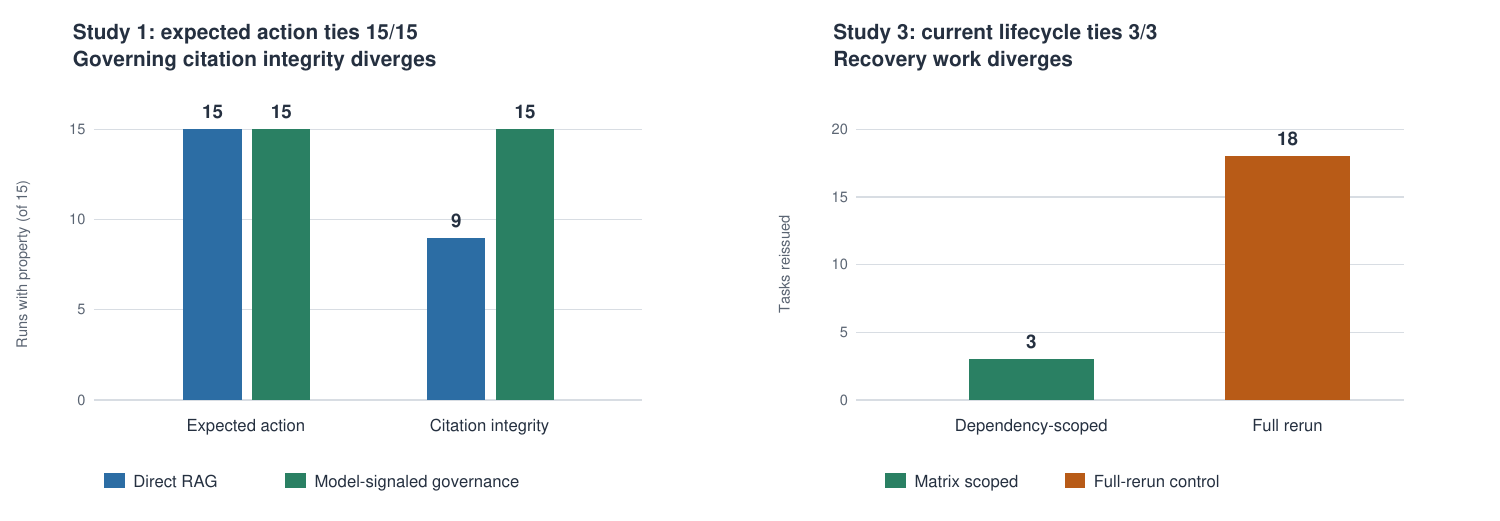}
\caption{Headline matched comparisons. Left: expected-action accuracy ties at
15/15, while runs without unsupported governing citations are 9/15 for direct
RAG and 15/15 for model-signaled governance. Right: both Study 3 arms reach
current state in 3/3 repeated seeds, while the full-rerun control reissues 18
tasks and dependency-scoped recovery reissues 3. Seeds are stability checks,
not independent task samples.}
\label{fig:headline-results}
\end{figure}

Figure~\ref{fig:headline-results} makes the two principal same-outcome
comparisons explicit. Study 1 ties on expected action while citation integrity
diverges; Study 3 ties on current lifecycle while reexecution scope diverges.

\subsection{Decision admission: the same outcome after explicit grounding repair}

L4 illustrates how the distinction enters a workflow. Phi omitted the required
preferred-vendor fact reference in the initial proposal on all three seeds. Both
initial proposals therefore failed the same contract. \matrixsys{} issued the
exact residual obligation, accepted one model-authored replacement only after
it passed revalidation, and admitted the corrected proposal 3/3. No task ran
before admission. The direct path retained the omission 3/3.

Both paths later reached the expected business state 3/3. The contribution is
not improved business accuracy: only the governed path repaired and recorded
the grounding before execution. A tested persistent-omission branch escalates
after the bounded rework allowance with zero task execution; the live Phi runs
exercised correction rather than escalation.

\subsection{Study 2: self-closure is not verified closure}

Table~\ref{tab:execution-provenance} reports C4 and C5. The shared Phi plan was
exact after generic self-review in every run. The direct condition nevertheless
accepted an unsupported DPA completion claim and self-closed 3/3 in both rungs.

\begin{table}[t]
\centering
\small
\begin{tabular}{p{0.12\linewidth} p{0.24\linewidth} p{0.43\linewidth} r}
\toprule
Rung & Direct control & \matrixsys{} & Checks \\
\midrule
C4 & Valid completion 0/3 after self-closing 3/3 & Returned
\texttt{rework\_required} and preserved the DPA residual 3/3 & 39/39 \\
C5 & Valid completion 0/3 after self-closing 3/3 & Retained the failed
attempt, directed one retry, independently verified it, and closed 3/3 & 45/45 \\
\bottomrule
\end{tabular}
\caption{Study 2: claimed versus verified completion. Checks are deterministic
lifecycle-action assertions, not independent samples.}
\label{tab:execution-provenance}
\end{table}

C4 deliberately stops at the residual obligation; evidence-backed completion
is zero in both conditions because no valid replacement is provided. C5 adds
that replacement. \matrixsys{} retains the invalid first receipt, issues the
specific evidence correction, validates the countersigned retry, and closes
from the new current receipt set. The direct condition reaches the same claimed
business outcome but never obtains valid DPA evidence.

The generic self-review control matters: the observed difference is not against
a one-shot model that was never asked to check its plan. The self-review
produced a complete plan, but it did not create an independent execution
verifier. A deterministic persistent-failure regression consumes the same
retry allowance, retains both invalid receipts, and escalates with the DPA
obligation unresolved.

\subsection{Study 3: currentness ties while recovery scope diverges}

C6 gives both conditions the same privacy amendment and makes both recovery
strategies safe. Table~\ref{tab:change-provenance} shows that both arms reached
the current lifecycle 3/3. \matrixsys{} traversed declared dependencies and
reissued DPA once per seed. The direct control conservatively repeated all six
tasks per seed.

\begin{table}[t]
\centering
\small
\begin{tabular}{lcc}
\toprule
Metric & \matrixsys{} & Full-rerun control \\
\midrule
Current lifecycle exact & 3/3 & 3/3 \\
Task reissues, total & 3 & 18 \\
Unaffected receipts preserved & 5 per seed & 0 per seed \\
Governed stale residue & 0 & -- \\
Independent supervisor closure & 3/3 & 0/3 \\
\bottomrule
\end{tabular}
\caption{Study 3: change recovery. The direct control is intentionally
conservative and current; the measured difference is dependency-scoped work.}
\label{tab:change-provenance}
\end{table}

All 48 governed lifecycle assertions passed. The result demonstrates a sixfold
difference in task reexecution on one declared amendment, not a general cost
ratio. Change detection itself was supplied by the fixture and is not measured.

\subsection{Study 4: contract fidelity with a bounded refusal guarantee}

Phi proposed escalation on none of the 12 structurally incomplete Experiment 6
executions. The public completeness invariant blocked each confident ruling and
made escalation effective in all 12. The same enforcement path later stayed
silent on every complete pilot execution.

\begin{table}[t]
\centering
\small
\begin{tabular}{p{0.25\linewidth} cc p{0.29\linewidth}}
\toprule
Packet set & Unique contracts & Executions & Result \\
\midrule
Structurally incomplete & 2 & 12 & Phi escalation 0/12; effective Matrix escalation 12/12 \\
Structurally complete & 12 & 120 & False escalations 0/120; specificity 1.00 \\
Resolvable complete subset & 10 & 100 & False escalations 0/100 \\
\bottomrule
\end{tabular}
\caption{Study 4: metadata-completeness invariant. Complete-contract specificity
was added post hoc on the same authored fixture.}
\label{tab:safety-invariant}
\end{table}

This result rules out the degenerate implementation that escalates every
packet, but only within the declared contract family. It demonstrates that the
code faithfully implements metadata-decidable refusal; it does not show that
the contract captures semantic completeness. The ordinary policy-adjudication
comparison was weak and remains outside the claim: graph 27/60 versus flat
21/60, descriptive $p=0.238$.

\subsection{Study 5: assurance survives while context accuracy ties}

C7's raw history exceeded the treatment budget on all three seeds
($6{,}160 > 4{,}096$ estimated tokens). \matrixsys{} composed dependency gating,
receipt rework, selective invalidation, and current supervisor closure 3/3; all
60 lifecycle assertions passed. The direct strict lifecycle remained 0/3
because its registry completion claim was never independently verified.

Crucially, fresh truth, Matrix scoped state, and the fair Phi-compacted control
all selected current closure 3/3. C7 therefore supports composition of the
assurance lifecycle under a bounded state surface, not a context-accuracy
advantage.

\subsection{Supporting state-surface stress test}

The earlier short-context frontier diagnostic reached 40/40 exact recovery in
both governed and transcript conditions. In Experiment 3, the same frontier
reviser was then held fixed while a two-correction history was forced through
compaction. Table~\ref{tab:compaction-support} reports the result.

\begin{table}[t]
\centering
\small
\begin{tabular}{lcccc}
\toprule
Compaction cell & Ledger exact & Transcript exact & Ledger F1 & Transcript F1 \\
\midrule
Frontier, 800 & 40/40 & 1/40 & 1.000 & 0.025 \\
Mini, 800 & 40/40 & 7/40 & 1.000 & 0.407 \\
Mini, 400 & 40/40 & 0/40 & 1.000 & 0.217 \\
\bottomrule
\end{tabular}
\caption{Supporting state-surface stress test: second-correction recovery. The
final reviser is unchanged across cells.}
\label{tab:compaction-support}
\end{table}

The compacted failures retained no complete answer-lineage items. A mechanical
scope query exposed six answer-lineage and ten non-answer-lineage references,
so the ledger was not filtered to the answer set. Faithful prose rendered from
the same query also achieved 40/40, identifying complete targeted provenance
retrieval rather than JSON syntax as the primary mechanism in this fixture.
The non-monotonic F1 values are set-overlap behavior rather than an exact-score
contradiction: the frontier summaries usually omitted the target lineage almost
entirely, while the mini 400-token summaries retained scattered target items but
never the complete required set.

This is supporting evidence that a governed state surface can become
outcome-relevant when provenance is dropped. It is not promoted to a robust
product accuracy claim: L6's later difference reduced to a top-$K$ recall
cutoff, and C7's compactor preserved current truth.

\subsection{Study 6: the completeness contract failed to transfer}

Experiment 7's reviewers agreed on structural completeness for 26/30 synthetic
packets (Cohen's $\kappa=0.659$). The frozen contract detected all 23
adjudicated-incomplete packets but blocked all seven reviewer-complete packets:
sensitivity 1.00, specificity 0.00, and accuracy 0.767. The validity gate
failed, so no downstream Experiment 7 workflow comparison is claim-bearing.

A final semantic-normalization diagnostic used a separately expanded, balanced
40-packet set assembled after the original 30-packet corpus was exposed. It
blocked all 20 reviewer-incomplete packets but admitted only 3/20
reviewer-complete packets, for specificity 0.15 and accuracy 0.575. Because
this was a burned-corpus repair diagnostic rather than a fresh transfer test,
it does not replace the controlling zero-specificity result.

The negative result separates mechanism correctness from contract validity.
The authored ladders show that \matrixsys{} can enforce a declared contract;
Experiment 7 shows that the contract's notion of completeness may not transfer
to independently produced synthetic packets.

\section{Discussion}

\subsection{The tied outcomes are the central result}

The most title-aligned comparisons are not accuracy wins. Direct RAG and
governed retrieval selected the same actions in Experiment 5. Both C6 recovery
strategies reached current state. L4 reached the same final business state in
both conditions. If evaluation stops at the action or final state, these cases
are ties.

The workflows differ on institutionally material questions: which evidence was
allowed to govern, whether completion was independently proven, and which prior
work a later change invalidated. The results therefore support measuring task
success and governance as separate axes. Governance is not a substitute for a
correct outcome; it is additional structure needed to justify, close, and
revise that outcome.

\subsection{One provenance concept across three lifecycle surfaces}

Decision, execution, and change provenance are not unrelated features. Each
binds a consequential state transition to the prior objects that justify it:

\begin{itemize}
  \item a decision binds to accepted authority and facts;
  \item a completion verdict binds to a task, definition of done, and current
  receipt evidence; and
  \item an invalidation or preservation decision binds to the versioned object
  and dependency path affected by a change.
\end{itemize}

This common representation enables one append-only lifecycle rather than
separate citation, workflow, and audit subsystems. It also supports residual
obligations: a failed decision, receipt, or recovery step remains explicit
instead of disappearing behind a later green status.

The synthesis, rather than any individual primitive, is the architectural
claim. Event logs provide replay, runtime monitors enforce declared rules,
citation systems bind claims to evidence, tool contracts gate effects, and
dependency graphs support selective recomputation. \matrixsys{} uses those
ideas to maintain one institutional lifecycle from versioned authority and
facts through admitted decisions, obligations, receipts, closure, and later
invalidation. The transfer failure is equally part of the contribution: a
faithful mechanism does not validate the institutional contract it executes.

\subsection{Why not just RAG?}

Experiment 5 answers a narrow version of this question. Retrieval made relevant
text available, and its recall was complete on the applicable-policy cases. It
did not by itself establish which retrieved records were institutionally
effective. Direct RAG could therefore choose the expected action while attaching
plausible but non-governing evidence.

The baseline is not a cite-everything strawman. It shares the model, physical
query, retriever, top $K$, records, authority metadata, action space, and
context budget, and it is explicitly instructed to reject the same distractor
classes. It is still a direct action-and-citation pipeline. A RAG system with an
independent applicability verifier may close the measured gap. If it also
records accepted authority, binds that state to downstream tasks, checks
receipts, and propagates later changes, it is implementing much of the
governance pattern studied here.

The governed path uses more inference and a separate adjudication call. The
current evidence cannot distinguish how much of Experiment 5 comes from
decomposition, prompt specialization, or persistent binding. It establishes the
system comparison, not an isolated ledger effect.

\subsection{Why not ask the model to check its work?}

C4--C6 use a stronger control than one-shot generation. The same Phi plan and
one explicit generic self-review are shared before the conditions diverge. The
self-review successfully produces exact obligation plans, yet the direct
condition still accepts its own unsupported completion claim. Planning review
and completion verification are different functions. This is consistent with
prior findings that intrinsic self-correction and self-verification can fail
without a sound external signal \citep{huang2024selfcorrect,
stechly2025selfverification}; the present result instantiates that distinction
at an evidence-backed workflow closure boundary.

Another model call could serve as a verifier, but independence then becomes a
system property: the verifier requires a frozen evidence contract, access to
external or tool-produced proof, an immutable record of failed attempts, and
authority to prevent closure. Those properties do not arise from reflective
language alone. The verifier may itself be a model where semantic judgment is
necessary; deterministic infrastructure can still own the surrounding identity,
retry, and transition rules.

\subsection{Hard mechanics around soft judgment}

Experiment 6 illustrates the useful boundary. Phi proposed escalation in 0/12
executions where a mechanically required authority class was missing. Once that
absence was represented as a public completeness contract, the framework could
guarantee refusal for the declared condition and remain silent on authored
complete contracts. No larger model is required to enforce a missing-key
invariant.

The framework cannot make the contract semantically complete. Experiment 7
shows why this distinction matters: perfect authored-fixture contract fidelity
coexisted with severe over-blocking after synthetic packet authorship and review
were separated from the contract. Determinism makes a rule reproducible; it does
not make the rule valid.

This suggests an engineering split. Metadata-decidable invariants can be hard,
model-independent gates. Applicability, completeness, and conflict adjudication
that require open-world interpretation remain soft judgments and may require
models or humans. The ordinary Experiment 6 reasoning comparison was weak,
which reinforces rather than undermines that boundary.

\subsection{Capability changes the reasoner, not the institutional requirement}

A stronger model may avoid some citation errors, remember a correction through
compaction, or catch an unsupported claim during self-review. The frequency of
intervention can therefore fall as capability improves. The institutional
requirements do not disappear: a reviewer still needs to know what authority
governed, what proof closed the task, and what changed after an amendment.

The supporting context experiments sharpen this point without supporting a
universal accuracy claim. A frontier model reconstructed short reliable history
perfectly, while mechanical provenance became load-bearing under one forced
compaction treatment. Later product fixtures did not reproduce a robust context
accuracy gap. The durable claim is therefore the inspectable state and lifecycle,
not permanent superiority over model memory.

\subsection{Governance is not objective truth}

\matrixsys{} records accepted institutional state under explicit authority and
evidence policies. It does not equate acceptance, consensus, or ledger presence
with objective truth. The mandatory-governance errors demonstrate that a bad
semantic choice can be promoted. The Experiment 7 transfer failure demonstrates
that an authored completeness contract can be wrong for new packets.

The value of the trace is that these judgments and dependencies are visible,
versioned, and revisable. When an executable or independently checkable oracle
exists, the supervisor can block false closure. Where no complete oracle exists,
governance improves accountability and controlled escalation but cannot
guarantee correctness.

\subsection{Product implication}

The product implied by these results is an institutional-integrity layer, not a
general accuracy enhancer. Its stable responsibilities are to distinguish
candidate evidence from admitted authority, compile accepted state into
obligations, require evidence for closure, preserve failures and residual work,
and target recovery after change. Models, prompts, retrievers, and execution
adapters can be replaced around that layer.

The next validation milestone is correspondingly concrete: an independently
authored enterprise task whose authority, facts, definition of done, execution
evidence, and review judgments are supplied by people who did not design the
contract. More self-authored rungs would expand coverage but not external
validity.

\section{Evidence Boundaries}

The results should be read as mechanism evidence, not as a deployment-rate
estimate. The experiments show that external governance can preserve provenance,
direct recovery, and make evidence grounding inspectable under controlled
conditions. They do not show that every enterprise workflow can be specified
correctly, that all dependency declarations will be complete, or that governed
agents will outperform ungoverned agents on every correctness metric.

Several boundaries are especially important. The main fixtures are authored and
synthetic, with seeds measuring output stability rather than independent task
designs. Some lifecycle counts are deterministic contract assertions rather
than population samples. Directed rework tests the governance loop after a
failure is detected; it does not prove that the model would autonomously
discover the missing fact or correction. The retrieval study compares governed
applicability adjudication against direct RAG, but does not establish
superiority over all possible citation-verifying RAG systems.

The paper also separates assurance from semantic capability. Matrix can make
state lineage, residual obligations, and closure criteria explicit, but it does
not make a weak model semantically equivalent to a stronger one. Later Phi
diagnostics support this boundary: verifier feedback can rescue localized
repair cases, while diffuse multi-constraint repairs and policy-modality
judgments remain model-capability limits.

Appendix~\ref{app:evidence-boundaries} gives the detailed evidence-boundary
record, including post-hoc framing controls, transfer-gate failures,
post-freeze verifier hardening, model/runtime coverage, and claims not
permitted by the current evidence.

\section{Related Work and Novelty Boundary}

The individual mechanisms in \matrixsys{} have substantial prior art. The
paper's claimed contribution is their integration into one governed-work
lifecycle and an evaluation that separates expected outcome from decision,
execution, and change provenance. Table~\ref{tab:nearest-neighbors} states that
boundary against the closest research lines.

\begin{table}[t]
\centering
\small
\begin{tabular}{p{0.18\linewidth} p{0.20\linewidth} p{0.20\linewidth} p{0.31\linewidth}}
\toprule
Research line & Primary object & Control boundary & Distinction studied here \\
\midrule
Citation grounding
& Claim or proposed action & Generation or pre-tool check
& Institutional applicability persists into tasks, closure, and later change. \\
Runtime enforcement
& Path, tool call, state, or durable effect & Before action or at commit
& One state model links authority and facts to decisions, obligations, receipts,
verdicts, and rework. \\
Durable workflows
& Checkpointed workflow state or artifact DAG & Resume, replay, or update
& Authority applicability and evidence-sufficient closure govern what may remain
current. \\
\matrixsys{} & Institutional work objects & Admission through post-change recovery
& Matched outcome/provenance evaluation plus an explicit contract-transfer
failure. \\
\bottomrule
\end{tabular}
\caption{Nearest-neighbor scope comparison. Rows describe each line's stated
center of gravity, not an assertion that other implementations cannot be
extended to the \matrixsys{} lifecycle.}
\label{tab:nearest-neighbors}
\end{table}

\paragraph{Citation, attribution, and decision provenance.}
RAG couples generation to retrieved non-parametric memory
\citep{lewis2020rag}. ALCE evaluates citation correctness and completeness in
retrieval-grounded generation \citep{gao2023alce}, while GAVEL binds atomic
subclaims to evidence units and performs deterministic citation validation
\citep{xu2026gavel}. Controlled authorization audits show that source authority
can change agent action even when proposition content is held fixed
\citep{liao2026provenance}; ProvenanceGuard checks whether a proposed tool call
is supported by traceable evidence before execution \citep{she2026guard}.
These are direct neighbors, not merely adjacent work. Study 1 is narrower than
state-of-the-art citation-verifying pipelines: its direct-RAG arm has no
independent applicability verifier. Its role is to demonstrate, on the authored
corpus, that expected action and governing citation can diverge. \matrixsys{}
then persists the admitted authority choice as input to downstream work.

\paragraph{Runtime governance, policy as code, and verified effects.}
Policy-as-code engines such as Open Policy Agent separate policy evaluation
from application enforcement over structured input \citep{opa2026}. AgentSpec
provides a language for runtime constraints with triggers, predicates, and
enforcement actions \citep{wang2025agentspec}; path-based governance treats the
partial execution path and proposed next action as the policy object
\citep{kaptein2026paths}. ToolGate maintains typed symbolic state and uses
preconditions and postconditions to gate tool invocation and state commit
\citep{liu2026toolgate}. Commit-time authorization is especially close to the
paper's title claim: visible endpoint success may remain high after the witness
authorizing a durable effect has become invalid \citep{santos2026committime}.
\matrixsys{} does not claim to originate deterministic runtime enforcement,
contract-gated tools, or the distinction between visible and authorized
success. Its different unit is an institutional workflow whose decision,
obligation, receipt, supervisor closure, and later rework share versioned
authority and fact dependencies.

\paragraph{Durable orchestration and change propagation.}
Current workflow systems already provide typed graphs, checkpointing, durable
resumption, replay, and forking. Microsoft Agent Framework supersedes the older
AutoGen-centered orchestration reference and supports graph workflows,
checkpointing, human interaction, and durable execution
\citep{microsoft2026agentframework}. LangGraph similarly persists per-step
checkpoints for replay, fault recovery, and time travel
\citep{langgraph2026persistence}. Execution Lineage represents AI-native work as
an artifact DAG with explicit dependencies and identity-based replay, and
demonstrates selective propagation with unaffected-artifact preservation
\citep{rosen2026lineage}. This directly overlaps Study 3. The append-only log,
dependency graph, and selective replay are therefore substrate rather than a
standalone novelty claim. The distinction tested here is their use with
versioned institutional authority, evidence contracts, supervisory closure, and
residual obligations in the same lifecycle.

\paragraph{Truth maintenance, data provenance, and process evidence.}
Truth maintenance systems record justifications and revise dependent beliefs
when assumptions change \citep{doyle1979tms,dekleer1986atms}; formal belief
revision studies rational knowledge-base change \citep{agm1985}. Database
provenance semirings explain how query outputs depend on input records
\citep{green2007semirings}. W3C PROV represents entities, activities, agents,
derivations, and responsibility \citep{moreau2013provdm}, while process mining
uses event logs for discovery and conformance checking
\citep{vanderaalst2012manifesto}. \matrixsys{} applies related dependency ideas
to a live agent-control surface. It does not propose a new general provenance
algebra or interoperability standard.

\paragraph{Auditability and specification validity.}
Runtime verification can faithfully enforce observed executions against a
specification \citep{leucker2009runtime}. Recent work frames agent auditability
across lifecycle coverage, policy checkability, recoverability, responsibility,
and evidence integrity \citep{nian2026auditable}. \matrixsys{} operationalizes
a subset of that agenda in one prototype. Study 6 adds a separate warning:
mechanism fidelity does not establish that an authored institutional contract
is semantically valid on newly produced packets. That negative transfer result,
and its role as a gate on downstream claims, is part of the paper's novelty
boundary rather than an implementation footnote.

\section{Future Validation and System Scope}

\subsection{Independently authored governed-work study}

The next claim-bearing study should begin with one real or realistic enterprise
task whose policy owners, packet authors, execution adapters, definitions of
done, and audit reviewers did not design \matrixsys{} or its contract. The
study should freeze:

\begin{itemize}
  \item the authority and fact inventory before workflow execution;
  \item task preconditions and definitions of done before evidence is observed;
  \item a direct self-review baseline, a citation-verifying retrieval baseline,
  and the governed condition under matched models and evidence;
  \item independent judgments of applicability, completion, and affected-work
  scope; and
  \item stop/go thresholds for false blocks, missed residual work, stale
  residue, and human review burden.
\end{itemize}

The target outcome is not another decision benchmark. It is one task that the
system can ground, issue, supervise, and close from independently verifiable
proof, with unresolved work preserved when closure is not justified.

\subsection{Contract validity and invariant families}

Experiment 7 shows that contract transfer is a first-class research problem.
Future work should separate policy authorship, schema translation, packet
production, and completeness review across independent humans or organizations.
Calibration and test packets must remain disjoint, and contract changes after
label access must trigger a new held-out evaluation.

The missing-authority rule is one member of a possible invariant family:
expired authority, superseded policy, delegation bounds, jurisdiction, and
required approval multiplicity may also be metadata-decidable. Each candidate
invariant requires both incomplete and complete held-out cases. A broad family
is valuable only if it avoids turning uncertainty into systematic over-blocking.

\subsection{Stronger matched baselines and operational measures}

Experiment 5 should be repeated against an equal-call citation-verifying RAG
pipeline. Completion studies should compare deterministic verification with a
separate model verifier that receives the same evidence contract and tool
results. Measurements should include latency, tokens, tool calls, human
intervention time, false escalation, rework attempts, and the cost of broad
versus selective recovery.

\subsection{Interoperable governed state}

A generalized runtime requires stable schemas for authority, facts, tasks,
receipts, verdicts, invalidations, and replacement lineage; identity and role
assignment; allowed tools; and pluggable outcome verifiers. Mapping these
records to W3C PROV and process-event standards would make traces portable to
existing audit and process-mining tools. The current prototype is a local
implementation, not a completed interoperability standard.

\subsection{Causal process intelligence}

Branch-native replay could eventually compare actual execution with a branch in
which a governing variable changes. Aggregated traces could expose process
variants, bottlenecks, and intervention candidates. Future-variable models might
support projective scenarios across cost, time, capacity, and risk. These are
research hypotheses, not capabilities established by this paper. A modeled
counterfactual is not proof of causality, and a projected branch is not a
guaranteed future.

\section{Conclusion}

Correct actions are necessary, but they do not establish governed execution.
An institutional workflow must also identify what authority and facts governed
the decision, what independent evidence proves completion, and how later
changes affected prior work.

The experiments make those differences observable. Direct RAG and governed
retrieval selected the same expected actions, yet differed in whether
non-governing evidence was cited. A generic model self-review produced complete
plans but did not prevent self-certified closure on invalid evidence; receipt
verification and bounded rework did. A conservative full rerun and
dependency-directed recovery both restored current state after an amendment,
but only the latter preserved unaffected proof and limited reexecution to
dependent work. A deterministic invariant forced refusal for a declared missing
authority class where the model never self-refused.

The negative boundary is equally important. Perfect implementation of an
authored completeness contract did not make that contract valid on
independently contextualized synthetic packets. \matrixsys{} can make declared
institutional rules inspectable and deterministic; it cannot manufacture
semantic truth or external validity.

The resulting claim is deliberately bounded: a deterministic causal-state layer
can preserve decision, execution, and change provenance around probabilistic
agents, allowing correct outcomes to be distinguished from outcomes that are
also grounded, verified, current, and replayable.

\bibliographystyle{plainnat}
\bibliography{references}

\begin{thebibliography}{25}
\providecommand{\natexlab}[1]{#1}
\providecommand{\url}[1]{\texttt{#1}}
\expandafter\ifx\csname urlstyle\endcsname\relax
  \providecommand{\doi}[1]{doi: #1}\else
  \providecommand{\doi}{doi: \begingroup \urlstyle{rm}\Url}\fi

\bibitem[Alchourr{\'o}n et~al.(1985)Alchourr{\'o}n, G{\"a}rdenfors, and
  Makinson]{agm1985}
Carlos~E. Alchourr{\'o}n, Peter G{\"a}rdenfors, and David Makinson.
\newblock On the logic of theory change: Partial meet contraction and revision
  functions.
\newblock \emph{The Journal of Symbolic Logic}, 50\penalty0 (2):\penalty0
  510--530, 1985.
\newblock \doi{10.2307/2274239}.

\bibitem[de~Kleer(1986)]{dekleer1986atms}
Johan de~Kleer.
\newblock An assumption-based {TMS}.
\newblock \emph{Artificial Intelligence}, 28\penalty0 (2):\penalty0 127--162,
  1986.
\newblock \doi{10.1016/0004-3702(86)90080-9}.

\bibitem[Doyle(1979)]{doyle1979tms}
Jon Doyle.
\newblock A truth maintenance system.
\newblock \emph{Artificial Intelligence}, 12\penalty0 (3):\penalty0 231--272,
  1979.
\newblock \doi{10.1016/0004-3702(79)90008-0}.

\bibitem[{European Union}(2024)]{eu2024aiact}
{European Union}.
\newblock Regulation ({EU}) 2024/1689 laying down harmonised rules on
  artificial intelligence, 2024.
\newblock URL \url{https://eur-lex.europa.eu/eli/reg/2024/1689/oj}.
\newblock Article 12.

\bibitem[Gao et~al.(2023)Gao, Yen, Yu, and Chen]{gao2023alce}
Tianyu Gao, Howard Yen, Jiatong Yu, and Danqi Chen.
\newblock Enabling large language models to generate text with citations.
\newblock In \emph{Proceedings of the 2023 Conference on Empirical Methods in
  Natural Language Processing}, pages 6465--6488. Association for Computational
  Linguistics, 2023.
\newblock \doi{10.18653/v1/2023.emnlp-main.398}.
\newblock URL \url{https://aclanthology.org/2023.emnlp-main.398/}.

\bibitem[Green et~al.(2007)Green, Karvounarakis, and
  Tannen]{green2007semirings}
Todd~J. Green, Grigoris Karvounarakis, and Val Tannen.
\newblock Provenance semirings.
\newblock In \emph{Proceedings of the Twenty-Sixth ACM SIGMOD-SIGACT-SIGART
  Symposium on Principles of Database Systems}, pages 31--40. ACM, 2007.
\newblock \doi{10.1145/1265530.1265535}.

\bibitem[Huang et~al.(2024)Huang, Chen, Mishra, Zheng, Yu, Song, and
  Zhou]{huang2024selfcorrect}
Jie Huang, Xinyun Chen, Swaroop Mishra, Huaixiu~Steven Zheng, Adams~Wei Yu,
  Xinying Song, and Denny Zhou.
\newblock Large language models cannot self-correct reasoning yet.
\newblock In \emph{International Conference on Learning Representations}, 2024.
\newblock URL \url{https://openreview.net/forum?id=IkmD3fKBPQ}.

\bibitem[{IEEE Task Force on Process Mining}(2012)]{vanderaalst2012manifesto}
{IEEE Task Force on Process Mining}.
\newblock Process mining manifesto.
\newblock In \emph{Business Process Management Workshops}, volume~99 of
  \emph{Lecture Notes in Business Information Processing}, pages 169--194.
  Springer, 2012.
\newblock \doi{10.1007/978-3-642-28108-2_19}.

\bibitem[Kaptein et~al.(2026)Kaptein, Khan, and Podstavnychy]{kaptein2026paths}
Maurits Kaptein, Vassilis-Javed Khan, and Andriy Podstavnychy.
\newblock Runtime governance for {AI} agents: Policies on paths, 2026.
\newblock URL \url{https://arxiv.org/abs/2603.16586}.

\bibitem[{LangChain}(2026)]{langgraph2026persistence}
{LangChain}.
\newblock {LangGraph} persistence, 2026.
\newblock URL
  \url{https://docs.langchain.com/oss/python/langgraph/persistence}.
\newblock Accessed 2026-08-11.

\bibitem[Leucker and Schallhart(2009)]{leucker2009runtime}
Martin Leucker and Christian Schallhart.
\newblock A brief account of runtime verification.
\newblock \emph{The Journal of Logic and Algebraic Programming}, 78\penalty0
  (5):\penalty0 293--303, 2009.
\newblock \doi{10.1016/j.jlap.2008.08.004}.

\bibitem[Lewis et~al.(2020)Lewis, Perez, Piktus, Petroni, Karpukhin, Goyal,
  K{\"u}ttler, Lewis, tau Yih, Rockt{\"a}schel, Riedel, and
  Kiela]{lewis2020rag}
Patrick Lewis, Ethan Perez, Aleksandra Piktus, Fabio Petroni, Vladimir
  Karpukhin, Naman Goyal, Heinrich K{\"u}ttler, Mike Lewis, Wen tau Yih, Tim
  Rockt{\"a}schel, Sebastian Riedel, and Douwe Kiela.
\newblock Retrieval-augmented generation for knowledge-intensive {NLP} tasks.
\newblock In \emph{Advances in Neural Information Processing Systems},
  volume~33, pages 9459--9474, 2020.
\newblock URL
  \url{https://proceedings.neurips.cc/paper/2020/hash/6b493230205f780e1bc26945df7481e5-Abstract.html}.

\bibitem[Liao(2026)]{liao2026provenance}
Junchi Liao.
\newblock Auditing provenance sensitivity in {LLM} agent action selection,
  2026.
\newblock URL \url{https://arxiv.org/abs/2607.20827}.

\bibitem[Liu et~al.(2026)Liu, Peng, Cao, Wang, Deng, Chen, Yin, and
  Zhang]{liu2026toolgate}
Yanming Liu, Xinyue Peng, Jiannan Cao, Xinyi Wang, Songhang Deng, Jintao Chen,
  Jianwei Yin, and Xuhong Zhang.
\newblock {ToolGate}: Contract-grounded and verified tool execution for {LLM}s.
\newblock In \emph{Findings of the Association for Computational Linguistics:
  ACL 2026}, pages 9653--9684. Association for Computational Linguistics, 2026.
\newblock \doi{10.18653/v1/2026.findings-acl.470}.
\newblock URL \url{https://aclanthology.org/2026.findings-acl.470/}.

\bibitem[{Microsoft}(2026)]{microsoft2026agentframework}
{Microsoft}.
\newblock Microsoft agent framework workflows, 2026.
\newblock URL
  \url{https://learn.microsoft.com/en-us/agent-framework/workflows/}.
\newblock Accessed 2026-08-11.

\bibitem[Moreau and Missier(2013)]{moreau2013provdm}
Luc Moreau and Paolo Missier.
\newblock {PROV-DM}: The {PROV} data model.
\newblock W3c recommendation, World Wide Web Consortium, 2013.
\newblock URL \url{https://www.w3.org/TR/prov-dm/}.

\bibitem[Nian et~al.(2026)Nian, Yuan, Zhang, Li, and Zhao]{nian2026auditable}
Yi~Nian, Aojie Yuan, Haiyue Zhang, Jiate Li, and Yue Zhao.
\newblock Auditable agents, 2026.
\newblock URL \url{https://arxiv.org/abs/2604.05485}.

\bibitem[{Open Policy Agent}(2026)]{opa2026}
{Open Policy Agent}.
\newblock Open policy agent documentation, 2026.
\newblock URL \url{https://www.openpolicyagent.org/docs}.
\newblock Accessed 2026-08-11.

\bibitem[Rosen and Rosen(2026)]{rosen2026lineage}
Josh Rosen and Seth Rosen.
\newblock From agent loops to deterministic graphs: Execution lineage for
  reproducible {AI}-native work, 2026.
\newblock URL \url{https://arxiv.org/abs/2605.06365}.

\bibitem[Santos-Grueiro(2026)]{santos2026committime}
Igor Santos-Grueiro.
\newblock Temporary authority, permanent effects: Commit-time authorization for
  {LLM} agents, 2026.
\newblock URL \url{https://arxiv.org/abs/2607.10487}.

\bibitem[She et~al.(2026)She, Liang, and Kang]{she2026guard}
Yining She, Yiliang Liang, and Eunsuk Kang.
\newblock Safeguarding {LLM} agents from misalignment through provenance
  analysis, 2026.
\newblock URL \url{https://arxiv.org/abs/2607.01236}.

\bibitem[Stechly et~al.(2025)Stechly, Valmeekam, and
  Kambhampati]{stechly2025selfverification}
Kaya Stechly, Karthik Valmeekam, and Subbarao Kambhampati.
\newblock On the self-verification limitations of large language models on
  reasoning and planning tasks.
\newblock In \emph{International Conference on Learning Representations}, 2025.
\newblock URL
  \url{https://proceedings.iclr.cc/paper_files/paper/2025/file/f3c5e56274140e0420baa3916c529210-Paper-Conference.pdf}.

\bibitem[Tabassi(2023)]{nist2023airmf}
Elham Tabassi.
\newblock Artificial intelligence risk management framework ({AI RMF} 1.0).
\newblock Technical Report NIST AI 100-1, National Institute of Standards and
  Technology, 2023.
\newblock URL \url{https://doi.org/10.6028/NIST.AI.100-1}.

\bibitem[Wang et~al.(2025)Wang, Poskitt, and Sun]{wang2025agentspec}
Haoyu Wang, Christopher~M. Poskitt, and Jun Sun.
\newblock {AgentSpec}: Customizable runtime enforcement for safe and reliable
  {LLM} agents, 2025.
\newblock URL \url{https://arxiv.org/abs/2503.18666}.

\bibitem[Xu et~al.(2026)Xu, Li, and Sheng]{xu2026gavel}
Ruoyu Xu, Gaoxiang Li, and Victor~S. Sheng.
\newblock {GAVEL}: Evidence-contract debate with mechanized scrutiny for
  provenance-grounded fact-checking.
\newblock In \emph{Findings of the Association for Computational Linguistics:
  ACL 2026}, pages 35907--35920. Association for Computational Linguistics,
  2026.
\newblock \doi{10.18653/v1/2026.findings-acl.1789}.
\newblock URL \url{https://aclanthology.org/2026.findings-acl.1789/}.

\end{thebibliography}

\appendix
\section{Detailed Evidence Boundaries}
\label{app:evidence-boundaries}

\paragraph{Post-hoc paper framing.}
The experiments were not preregistered as one three-provenance study. The paper
recomposes existing frozen artifacts after their results were known. To reduce
selection drift, the claim-evidence map was frozen before the rewrite, original
conditions and negative results are retained, and no cross-study aggregate is
reported. This discipline improves transparency but does not remove the risk of
post-hoc narrative selection.

\paragraph{Self-authored fixtures and small independent sample.}
Experiment 5 has five authored scenarios; C4--C7 and L4 each have one authored
case repeated over three seeds; Experiment 6 has 12 authored standard contracts
and two incomplete control types. Seeds measure output stability, not new task
designs. Counts such as 39/39 or 60/60 are deterministic lifecycle assertions,
not independent observations and not confidence about deployment rates.

\paragraph{Contract fidelity is partly by construction.}
Receipt gates, dependency traversal, and metadata-completeness checks are
designed to implement declared contracts. Perfect action assertions show that
the code follows those specifications. They do not independently establish that
the definitions of done, dependencies, or completeness requirements are
semantically correct. This is the intended mechanism claim, but it is weaker
than enterprise validity.

\paragraph{The transfer gate failed.}
Experiment 7 directly challenges contract validity. Its frozen completeness
contract had specificity 0.00 on seven reviewer-complete synthetic packets, and
the final burned-corpus normalization diagnostic reached only 0.15. The
reviewers were fresh role contexts, not independent humans, but the result is
still sufficient to reject a general semantic-completeness claim from the
authored ladders. Downstream Experiment 7 comparisons are diagnostic only.

\paragraph{Direct RAG uses less inference.}
Experiment 5 matches model, query, retriever, top $K$, records, metadata,
action contract, and context budget, but governed paths add applicability
adjudication. Direct RAG has no independent citation verifier. The study cannot
attribute the full difference to persistent state rather than decomposition or
additional inference, and it does not establish superiority over
citation-verifying RAG.

\paragraph{The self-review control is strong but not exhaustive.}
C4--C7 give the direct condition one generic review pass and the same revised
plan. Other prompting, verifier models, external workflow engines, or
tool-enforced completion contracts could reproduce parts of the behavior. Such
systems would be alternative implementations of the assurance functions, not
evidence that those functions are unnecessary.

\paragraph{Directed rework tests the loop.}
L4 and C5 issue exact residual obligations. They test detection, directed
correction, independent revalidation, and bounded escalation; they do not test
whether the model autonomously discovers the missing fact or evidence. The
model authors the replacement, but the framework specifies what the failed
contract requires.

\paragraph{Post-freeze verifier hardening.}
A later non-claim-bearing diagnostic on a separate experimental closure surface
false-closed two of three deceptive executions because a revised model response
bypassed final deterministic admission. The frozen C4--C7 fixtures use a
separate completion path that derives closure directly from the current receipt
set; they do not route a revised model verdict through the defective admission
branch, and current regression tests exercise that receipt-derived path. The
defect therefore does not alter their frozen counts, but it exposes an
implementation risk beyond those fixtures. The current kernel recomputes revised
receipt sets from check state and rejects contradictory verdicts; a bounded
follow-on diagnostic produced no verifier false close on two unsafe designs and
no false block on two complete endpoint designs. These are regression checks,
not population safety estimates or a live-model replication of C4--C7.

\paragraph{Synthetic execution and no human-load measurement.}
The completion ladders use controlled execution adapters and declared evidence
artifacts rather than production enterprise systems. Latency, human review
time, escalation burden, and monetary cost are not measured. Experiment 7's
review-time values are model-role workload proxies and are not used here.

\paragraph{Change detection is supplied.}
C6 measures invalidation and recovery after an observed amendment. It does not
measure whether Matrix, a retriever, or a human detects the change in the first
place, nor whether real dependency declarations are complete. Missing edges can
leave stale residue; overly broad edges can erase the measured churn benefit.

\paragraph{Safety specificity is post hoc and authored.}
Experiment 6 added complete contracts after observing the 12/12 sensitivity
result. The 120 complete executions repeat 12 authored contracts across seeds
and two surfaces. Zero false fires rule out escalate-all on that fixture, not a
population false-positive rate. The invariant covers metadata absence, not
semantic or open-world incompleteness.

\paragraph{Model and runtime coverage.}
The lifecycle ladders primarily use one quantized local Phi-4 checkpoint. The
retrieval and context studies use additional local and frontier models under
different runtimes and sampling controls. The paper does not estimate model-
family interactions or a scaling curve.

\paragraph{Context-accuracy non-replication.}
Experiment 3 demonstrates a controlled compaction mechanism, but later product
diagnostics did not establish a robust accuracy advantage. L6 reduced to a
top-$K$ recall cutoff, and C7's Phi compactor preserved current truth 3/3. The
paper therefore uses Experiment 3 only as supporting state-surface evidence.

\paragraph{No truth, enterprise, or forecasting guarantee.}
Ledger acceptance does not establish objective truth, and the current studies
do not estimate real enterprise prevalence or effectiveness. Counterfactual
forks, process optimization, and forecasting remain unvalidated research
directions.

\end{document}